\documentclass[journal]{IEEEtran}

\usepackage{graphicx}
\usepackage{cite}
\usepackage{picinpar}
\usepackage{amsmath}
\usepackage{url}
\usepackage[latin1]{inputenc}
\usepackage{colortbl}
\usepackage{soul}
\usepackage{multirow}
\usepackage{pifont}
\usepackage{color}
\usepackage{alltt}
\usepackage[hidelinks]{hyperref}
\usepackage{enumerate}
\usepackage{siunitx}
\usepackage{breakurl}
\usepackage{epstopdf}
\usepackage{pbox}
\usepackage{cite}
\usepackage{amsmath,amssymb,amsfonts}
\usepackage{mathrsfs}

\usepackage{algorithm}
\usepackage{algorithmic}
\usepackage{multirow}
\usepackage{graphicx}
\usepackage{textcomp}
\usepackage{xcolor}
\usepackage{colortbl}
\usepackage{threeparttable}
\usepackage{diagbox} 
\usepackage{tabu}
\usepackage{makecell}

\begin{document}
%
\title{A Hybrid Decomposition-based Multi-objective Evolutionary Algorithm for the Multi-Point Dynamic Aggregation Problem}
%
%
%

\author{
	Guanqiang~Gao,
	Bin~Xin,~\IEEEmembership{Member,~IEEE, }
	Yi~Mei,~\IEEEmembership{Senior Member,~IEEE,}
	Shuxin~Ding,
	and Juan Li,
	\thanks{
		This work was supported in part by the National Outstanding Youth
		Talents Support Program 61822304, in part by the National Natural Science
		Foundation of China under Grant 61673058, in part by the NSFC-Zhejiang
		Joint Fund for the Integration of Industrialisation and Information under
		Grant U1609214, in part by the Projects of Major International (Regional)
		Joint Research Program of NSFC under Grant 61720106011, and in part by
		Consulting Research Project of the Chinese Academy of Engineering (2019-
		XZ-7). (Corresponding author:
		Bin Xin.)
		
		Guanqiang Gao and Bin Xin are with the School of Automation, Beijing Institute of Technology, Beijing 100081, China; 
		Yi Mei is with School of Engineering and Computer Science, Victoria University of Wellington, PO Box 600, Wellington 6140, New Zealand.
	}
}

\markboth{Journal of \LaTeX\ Class Files,~Vol.~14, No.~8, August~2015}%
{Shell \MakeLowercase{\textit{et al.}}: Bare Demo of IEEEtran.cls for IEEE Journals}
%


\maketitle

\begin{abstract}
	An emerging optimisation problem from the real-world applications, named the multi-point dynamic aggregation (MPDA) problem, has become one of the active research topics of the multi-robot system. This paper focuses on a multi-objective MPDA problem which is to design an execution plan of the robots to minimise the number of robots and the maximal completion time of all the tasks. The strongly-coupled relationships among robots and tasks, the redundancy of the MPDA encoding, and the variable-size decision space of the MO-MPDA problem posed extra challenges for addressing the problem effectively. To address the above issues, we develop a hybrid decomposition-based multi-objective evolutionary algorithm (HDMOEA) using $ \varepsilon $-constraint method. It selects the maximal completion time of all tasks as the main objective, and converted the other objective into constraints.  HDMOEA 
decomposes a MO-MPDA problem into a series of scalar constrained optimization subproblems by assigning each subproblem with an upper bound robot number. 
All the subproblems are optimized simultaneously with the transferring knowledge from other subproblems. Besides, we develop a hybrid population initialisation mechanism to enhance the quality of initial solutions, and a reproduction mechanism to transmit effective information and tackle the encoding redundancy. Experimental results show that the proposed HDMOEA method significantly outperforms the state-of-the-art methods in terms of several most-used metrics.
\end{abstract}


%
\IEEEpeerreviewmaketitle

\section{Introduction}
{\color{blue}What is the MDPA problem, why is the MPDA problem is important}
The Multi-Point Dynamic Aggregation (MPDA) problem is a task planning problem of the multi-robot system, which comes from the real world \cite{skorobogatov2020multiple,xin2016coordinated, gao2020memetic, lu2018multi,lan2020survey}.
Recently, it has become one of the active research topics due to its applications such as bushfire elimination routing, search and rescue, and medical resource scheduling domains \cite{xin2017distributed,chen2020unifying,chen2020Multiagent}.
Unlike the majority of scheduling and routing problems \cite{zhang2020FS,pezzella2008genetic,toth2014vehicle,golden2008vehicle}, the demand of each task in the MPDA problem increase over time, Besides, all the robots can execute one task simultaneously to promote the efficacy of completing the task.  The traditional objective of the MPDA problem is to design execution plans for all the robots to execute geographically distributed tasks in order to minimise the makespan (the completion time of the last completed task). 
 
 {\color{blue}The previous researchers did not consider the MO-MPDA problem, we focus on the MO-MPDA problem}
 The previous researches about the MDPA problem only focused on the single objective problem \cite{Duxin2018,xin2018estimation,teng2014market}. However, when the MPDA problem is applied to solve the real-world applications,  it always has two or more objectives that are usually conflicted. Thus, in this paper, we focus on the multi-objective MPDA (MO-MPDA) problem with two objectives. In the MO-MPDA problem,  there is a number of tasks (e.g. fire points) with time-varying demands in the workspace. One objective is to minimise the cost of using robots, and the cost is considered as the number of the robots in this paper.  The other objective is to complete all the tasks as soon as possible, and it is considered as the maximal completion time of all the tasks (makespan).

{\color{blue} There are some challenges in the MO-MPDA problem. }First, due to the multi-objective characteristic, the previous single-objective MPDA optimisation approaches \cite{teng2014market,gao2020ACO, gao2020memetic} cannot efficiently handle the two conflict objectives. When the number of used robots is changed, the whole single-objective approach needs to be adjusted to satisfy the changing robot number. Second, in the MO-MPDA problem, the length of the routing plan for one robot is not fixed, and the number of used robots is not fixed. Compared with the traditional multiobjective optimization problems with the fixed-length decision variables,  the MO-MPDA is difficult to solve due to their variable-length Pareto structure, where the number of variables in two different Pareto solutions might not be identical in the decision space. Third, the used robots are homogeneous so that there are many isomorphic execution plans for these robots. The fitness landscape of the MO-MPDA problem is so flat that some changes on the execution plan may not have ant effect on the objective value.  Last but not least, the time-varying demand and collaboration between robots lead the MO-MPDA problem has a large search space, in which an effective solution is non-trivial to be found.

{\color{blue} Introduce the MOEA/D method}. MOEA/D was first introduced in 2007 \cite{zhang2007moea}, and it shown very promising results for approximating the Pareto front.  
Decomposition is the main idea for  the MOEA/D method, it decomposes a multi-objective problem into a number of scalar optimisation subproblems.
Each subproblem learning valid information from its neighbouring subproblems so that all the subproblems are evolved collaboratively. 
Thank to the decomposition mechanism, the MOEA/D method has a relative low computational cost. Meanwhile, the MOEA/D method has a good population diversity due to the 
a scalar of the weight vectors in the objective space. There have been a variety of MOEA/D method with some promising results \cite{li2008multiobjective,li2020dmaoea,asafuddoula2014decomposition}. To the best of our knowledge, there is no previous work that has been applied MOEA/D to the MO-MPDA problem.

{\color{blue}Introduce the DMOEA-$ \varepsilon $C method.} 
The $ \varepsilon$-constraint method is firstly proposed in \cite{haimes1971bicriterion}, which selects one objective as the main objective and converts the other objectives into constraints \cite{deb2000efficient}.  The decomposition-based multi-objective evolutionary algorithm with the $ \varepsilon $-constraint framework (DMOEA-$ \varepsilon $C) \cite{chen2017dmoea}  firstly combined the $ \varepsilon$-constraint method and MOEA/D method to address the multi-objective problem. In DMOEA-$ \varepsilon $C, 
a multi-objective problem is decomposed into several constrained optimization subproblems with different upper bounds. These constrained optimization subproblems are optimised collaboratively using the neighbour information. DMOEA-$ \varepsilon $C is a very competitive MOEA method, and it has shown obvious advantages over MOEA/D on combinational optimisation problem \cite{chen2017dmoea,li2019improved,li2020dmaoea,li2020decomposition}.  Since the MPDA problem is a constraint combinational optimisation problem, we expect  DMOEA-$ \varepsilon $C to be effective in addressing the MO-MPDA problem.

{\color{blue} When we apply DMOEA-$\varepsilon $  to address the MO-MPDA problem, some issues need to be solved.} First, there is no existing mathematical model which can characterise the MO-MPDA problem. Second, the traditional solution-to-subproblem mechanism and subproblem-to-solution mechanism are difficult to be used in the MO-MPDA problem. Third, the random initialisation mechanism cannot generate high-quality solutions, even cannot generate feasible solutions for some tight-constraints scenarios.
Finally, it is challenging to tackle the encoding redundancy caused by homogeneous robots.

To address the above issues, we establish a MO-MPDA model and design a hybrid DMOEA-$ \varepsilon $C  (HDMOEA-$ \varepsilon $C) method.
Specifically, the contributions of this paper are shown as follows.

\begin{itemize}
	\item We formulate a MO-MPDA problem with two correlated objectives. One of the objectives is to minimise the cost of using robots (	
	the number of robots), and the other objective is to minimise the maximum completion time of all the tasks.
	\item According to the characteristics of the MO-MPDA problem, we design a novel HDMOEA-$ \varepsilon $C method including the designed solution-to-subproblem mechanism and subproblem-to-solution mechanism. 
	\item A hybrid initialisation strategy of the proposed  HDMOEA-$ \varepsilon $C method is designed to produce a fraction of high-quality initial solutions. In the initialisation strategy, a heuristic method considering the travelling cost, completion time, and the incremental rate of tasks generate a solution for each number of robots, the other solutions are generated by a random method.
	\item 
	In the designed reproduction strategy used in the reproduction process of HDMOEA-$ \varepsilon $C, all visiting sequences of all robots are first sorted according to the number of visiting tasks and the index of tasks. Then, a mutation operator and three crossover operators are used to reproduction new solutions in order to effective information transmission.
\end{itemize}

The rest of this paper is organised as follows. 
The related work and mathematical model of the MO-MPDA problem are presented in Section II. Then, Section III proposed the HDMOEA-$ \varepsilon $C method.  Section IV  presents the experimental results and performance analysis of the proposed mechanisms.
Finally, this paper is concluded in Section V.

\section{Background}
In this section, the related work about MPDA and MOEA is introduced first, and then the problem formulation of the MO-MPDA problem is given.

\subsection{Related Work}
\subsubsection{Multi-Point Dynamic Aggregation}

{\color{blue}Introduce the MPDA problem}
The MPDA problem is a novel optimisation problem that originates from the multi-robot system, and its objective is to design an optimal execution plan to maximise the 
performance of the multi-robot system \cite{liao2011multi, robin2016multi, xin2016coordinated,zhang2020review}. 
In the MPDA problem, each task has a time-varying demand, which grows over time. A set of robots is located at a depot and sent to execute these tasks collaboratively.
The MPDA problem can be applied to many real applications with time-varying demands and coordination behaviours (e.g. fire-fighting and area search). 
Due to the complex relationships among robots and tasks caused by the time-varying demand and coordinated execution, the MPDA problem is a very challenging and interesting problem. Over the past few years, researchers proposed several methods to address the MPDA problem \cite{mohandes2014motion,Duxin2018,gao2020memetic}.

{\color{blue}There are several heuristic approaches proposed to address the MPDA problem.}
A real fire-fighting system with a UAV and a UGV was implemented in \cite{ mohandes2014motion} by a simple heuristic approaches. The UAV in the fire-fighting system was used for the detection to reduce the uncertain of tasks, and the UGV was used to execute and complete the tasks.
Teng \textit{et al.} \cite{teng2014market} modified the bid value based-on the auction mechanism of the multi-robot system, considering robots' characteristics and the task's completion requirement. Du \textit{et al.} \cite{Duxin2018} proposed a recruitment strategy to address the MPDA problem. In the proposed recruitment method, when a robot is not able to complete its executing task, the robots will recruit some robots to completion the task together. Chen \textit{et al.} \cite{chen2020Multiagent} proposed a task assignment approach based on the completion time of each task. First, the task assignment approach tries to ensure every task is assigned to enough robots to complete it. Then, the approach assigns robots to the task with the largest reduction of the total completion time. 

{\color{blue} There are several meta-heuristic approaches proposed to address the MPDA problem.}
Although an execution plan of all the robots can be obtained by the aforementioned heuristic methods, the plan is not of a high quality due to the NP-hardness characteristic of the MPDA problem. Hence, several meta-heuristic approaches were proposed by researchers \cite{poggenpohl2012optimizing,hao2018hybrid,xin2018estimation}. 
A task planning approach hybridised with a greedy method and a simulated annealing algorithm is proposed by 
Poggenpohl \textit{et al.} \cite{poggenpohl2012optimizing}. The experimental results showed that the hybrid approach can lead to a better makespan and a better balance among the workloads of the robots. 
Another hybrid method combining differential evolution and estimation of distribution algorithm was proposed to address the MPDA planning problem by \cite{hao2018hybrid}. The experimental results showed that the hybrid method \cite{hao2018hybrid} outperforms the differential evolution in terms of the convergence speed and solution quality. A multi-model estimation of distribution algorithm with a multi-permutation encoding method was designed in \cite{xin2018estimation}. 
The experimental results showed that the method \cite{xin2018estimation} outperforms the genetic algorithm and random search method.
An adaptive coordination ant colony optimisation with an individual learning strategy is proposed to address the MPDA problem \cite{gao2020ACO}.
Experiments showed that the proposed ant colony optimisation algorithm significantly achieved a better performance than the state-of-the-art algorithms.

\subsubsection{Decomposition based  Multi-objective Evolutionary Algorithms}

Decomposition based MOEA has attracted great attention in recent years thanks to its decomposition and parallelism characteristics, and has become one of the mainstream methods to solve a multi-objective problem \cite{Li2014An,trivedi2016survey,xu2020survey}. MOEA/D \cite{zhang2007moea}  is the most popular one. MOEA/D decomposes a multi-objective optimisation problem into a series of single objective subproblems by using scalar aggregation functions (such as weighted sum, Tchebycheff and PBI) and a set of uniformly distributed weights, and uses EA to optimize all subproblems. MOEA/D defines the neighborhood relationship between subproblems according to the Euclidean distance between weight vectors. When solving each subproblem, only the information of its neighborhood subproblem is used, which makes MOEA/D have low computational complexity. Besides, MOEA/D implicitly guarantees the diversity of the population in the target space by giving a group of well-distributed weights. In recent years, many scholars have improved MOEA/D, mainly including the following aspects: weight design and adjustment, adding new search engines to the original MOEA/D framework, proposing new aggregation functions, and using MOEA/D and its improved forms to solve various types of real-world problems.

\begin{figure}[h]
	\vspace{-4mm}
	\centering  
	\includegraphics[width=6cm]{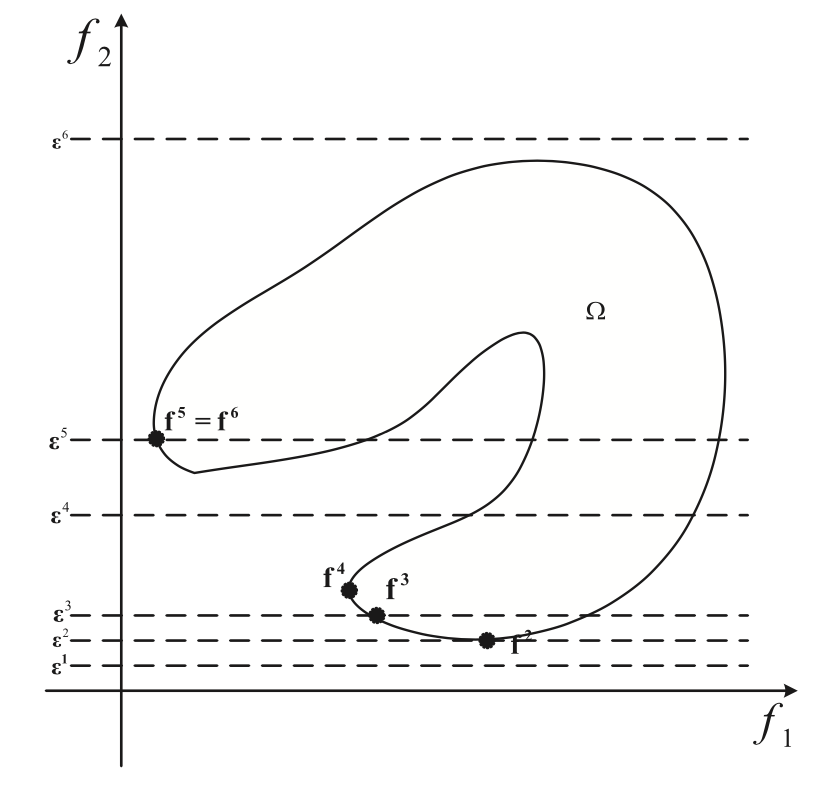}
	\caption{An example of the $ \varepsilon $-constraint method with different upper bounds.}
	\label{example_epsilon}
\end{figure}

Scholars have proposed variants of MOEA/D to promote performance. MOEA/D with dynamic resource allocation \cite{zhang2009performance} adjusts the computing resources allocated to each subproblem dynamically. MOEA/D with adaptive weight adjustment \cite{qi2014moeadawa} proposes a new weight generation method based on the analysis of the geometric relationship between the weight vector of Tchebycheff scalar function and the optimal solution of a corresponding scalar function.  
Ishibuchi \textit{et al}. [45]  demonstrated that employing a local replacement neighborhood structure is very important for the performance of MOEA/D. 
Wang \textit{et al.} proposed a global replacement scheme that assigns a new solution to its most suitable subproblems. 
Meanwhile, in the method, a dynamic adjusting replacement method of the neighborhood size was developed to promote the efficacy.
Zhou \textit{et al.} \cite{Moubayed2014D} extended the dynamic resource adjustment strategy in\cite{zhang2009performance}  to generalized resource allocation. The method \cite{Moubayed2014D}  allocates an improved probability vector for each subproblem, and allocates computing resources for each subproblem according to the improved probability vector. Chen \textit{et al.} proposed MOEA/D which decomposes a multi-objective problem to several $\varepsilon $-constraint optimization subproblem, which of the details are shown as follows.

\begin{equation}
\label{esop}
\begin{array}{l}
\begin{array}{*{20}{c}}
{{\mathop{\rm minimize}\nolimits} }&{{f_{main}} = {f_s}({\bf{x}})}
\end{array} + \rho \sum\limits_{i = 1}^m {{f_i}({\bf{x}})} \\
{\rm{~~subject~to}}  \left\{ \begin{array}{l}
\frac{{{f_i}({\bf{x}}) - z_i^*}}{{z_i^{nad} - z_i^*}} \le {\varepsilon _i},\forall i \in \{ 1,2, \cdots ,m\} /\{ s\} \\
{\bf{x}} = ({x_1},{x_2},...,{x_n}) \in \Omega
\end{array} \right.
\end{array}
\end{equation}
where $ s $ represents the predefined main objective index,  $0 \le{\bf{\varepsilon }}=({\varepsilon _1},...,{\varepsilon _{s - 1}},{\varepsilon _{s + 1}},...,{\varepsilon _m})\le 1$ is the upper bound vector, $\rho$ is a small positive number, ${{\bf{z}}^{\bf{*}}} = (z_1^*,...,z_m^*)$ and ${{\bf{z}}^{{\bf{nad}}}} = (z_1^{nad},...,z_m^{nad})$are the ideal point and the nadir point, respectively. An example of the $ \varepsilon $-constraint method with different upper bounds is shown in Fig.~\ref{example_epsilon}.  $ f_1 $ in Fig.~\ref{example_epsilon} is chosen as the main objective to be minimised, and a series of  upper bounds is applied into $ f_2 $. The black points represent the Pareto optimal solution within the given upper bounds vectors.

\subsection{Problem Formulation}
There is a number of tasks (e.g. fire points) with time-varying demands in the MO-MDPA problem. The decision maker wants to design a execution plan which completes all the tasks as soon as possible with as few robots as possible
We use an undirected graph $ G(\mathcal{V,E})$ to define the MO-MPDA problem. 
In the set of vertexes $\mathcal{V}= \{v_0, v_1,\dots,v_N\}$ of $ G $, $ v_0$ indicates the depot, and  
$ \{v_1,\dots, v_N\} $ indicates the set of tasks. Each task $ v_i $  has an inherent time-varying demand $ q_i(t) $, which is changed based on the following equation
\begin{equation}
q_i(t) =  q_i(0) + \alpha_i\times t, 
\end{equation}
where $ \alpha_i $ represents the inherent increment rate of task $v_i$.
In the set of edges of $ G $, every edge indicates a route among two tasks with the travel time $ t_{ij} $. 

\begin{figure}[!tbh]
	\vspace{-4mm}
	\centering  
	\includegraphics[width=6cm]{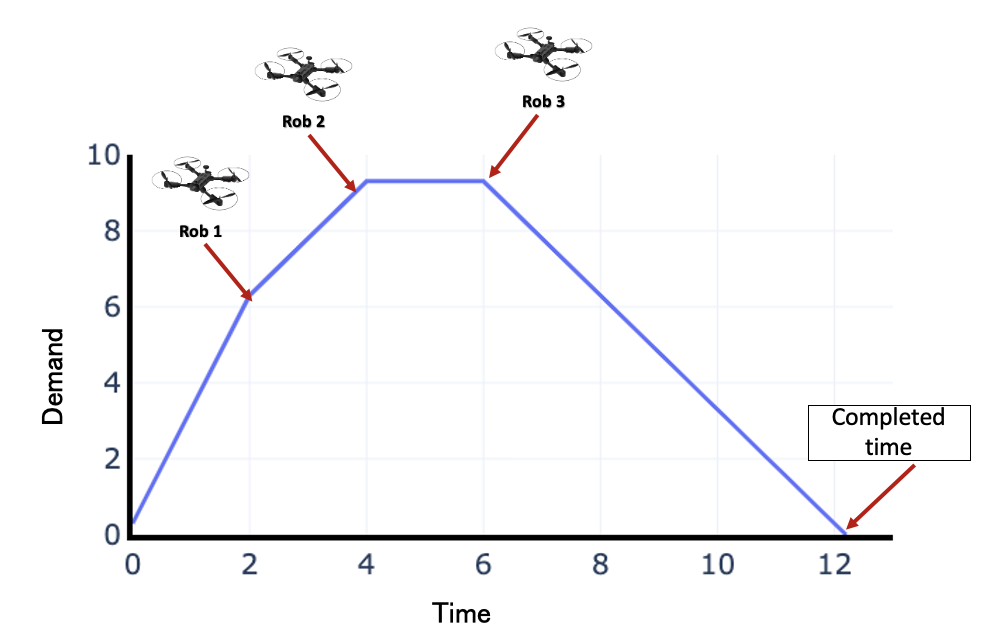}
	\caption{An example of the demand of an executed task over time.}
	\label{rateExample}
\end{figure}

Several homogeneous robots located at the depot are going to execute all the tasks. Every robot has the same ability $ \beta $, representing the amount of demand which it can reduce per time unit. Fig.~\ref{rateExample} shows an example demonstrating the relationship between the task demand and abilities of robots. In Fig.~\ref{rateExample}, the inherent increment rate $ \alpha_i $ of the exampled task is 3, and the ability of a robot is 1.5.
$ rob_1 $, $ rob_2 $ and $ rob_3 $ arrive at the task at time 2, 4 and 6 respectively. After $ rob_1 $ arrives, the current incremental rate of the task is 3 - 1.5 = 1.5.
After $ rob_2 $ reaches the task, the demand of the task is constant.  At time 6, $ rob_3 $ arrives, and the total robot ability is greater than the inherent increment rate.
Finally, the demand is decreased to 0, and the task is completed by these three robots.

Based on the aforementioned notations,  the MO-MPDA model can be defined as follows. 

\begin{equation}
\small
f_1 = \min \max_{i \in \{1,\cdots,N\}} ct_i \label{eq:obj1}
\end{equation}
\begin{equation}
\small
f_2 = \min m \label{eq:obj2}
\end{equation}
\begin{equation}
\small
s.t.  \sum_{j = 0} ^ N x_{ij}^k = \sum_{j = 0} ^ N x_{ji}^k,   \forall  i= 1,\dots,N,   \forall k =1,\dots,m 	\label{eq:sym}
\end{equation}
\begin{equation}
\small
\sum_{i = 0} ^ N  \sum_{k = 1} ^m x_{ij}^k \geq 1,  \forall j =1,\dots,N	\label{eq:exe} 
\end{equation}
\begin{equation}
\small
\sum_{i = 0} ^ N x_{ij}^k \leq 1, \forall j =1,\dots,N,  \forall k =1,\cdots,m \label{eq:once}
\end{equation}

\begin{equation}
\small
at_{0k}  = ct_{0} = 0, \forall k =1,\cdots,m \label{eq:depottime}
\end{equation} 
\begin{equation}
\small
at_{jk} = \sum_{i = 0}^N (ct_{i} + t_{ij}) x_{ij}^k, 	 \forall j =1,\dots,N,  \forall  k =1,\cdots,m	 \label{eq:ats}	
\end{equation}

\begin{equation}
\small
q_j(0) + \alpha_j ct_j  = \sum_{i=0} ^N \sum_{k = 1}^{m}  x_{ij}^k \beta  (ct_j - at_{jk}),
\forall j = 1,\dots,N \label{eq:dem}
\end{equation}
\begin{equation}
\small
\alpha_j < \sum_{k = 1}^m \sum_{i=0} ^N x_{ij}^k \beta, \forall  \label{eq:fea}
j =1,\dots,N
\end{equation}
\begin{equation}
\small
x_{ij}^k \in \{0, 1\}, i \neq j, \forall  i,j =1,\dots,N,   \forall  k =1,\dots,m \label{eq:domain}
\end{equation}
where the integer decision variable $ m $ represents the number of robots executing all the tasks, and the binary decision variables $x_{ij}^k $ takes 1 if $ rob_k $ goes from $ v_i $ to $ v_j $, and 0 otherwise.
$ at_{jk} $ represents the arrival time 
of $ rob_k $ at $ v_j $, and $ ct_j $ represents the completion time of $ v_j $. 
The notations used in the problem formulation are summarised in Table \ref{T1}.

One objective  (\ref{eq:obj1}) is to minimise the maximum completion time of all the tasks, and the other objective  (\ref{eq:obj2}) is minimise the number of robots executing all the tasks.
Constraint (\ref{eq:sym}) indicates that the number of outgoing routes equals the number of incoming routes for each task and each robot.
Constraint (\ref{eq:exe}) ensures that each task is executed by at least one robot. 
Constraint (\ref{eq:once}) ensures that each task is executed by each robot at most once.	Constraint (\ref{eq:depottime}) sets the arrival time and completion time for the depot to 0. Constraint (\ref{eq:ats}) specifies the relationship between $ct_j$, $at_{jk}$ and $x^k_{ij}$. 
Constraint (\ref{eq:dem}) implies that a task is completed when its demand decreases to zero (i.e. the accumulated demand from time 0 to $ct_j$ equals the total demand reduced by the robots executing the task during this time period). It also shows the time-varying characteristic of the task demand. Constraint (\ref{eq:fea}) indicates that for each task, the total ability of the robots executing it must be greater than its inherent increment rate. Otherwise, the task can never be completed. Constraint (\ref{eq:domain}) sets the binary domain of the decision variables.


\begin{table}[tb]
	\centering
	\footnotesize
	\caption{The notations used in the paper}
	\label{T1}
	\begin{tabular}{ c| c }
		\hline
		\hline
		\textbf{Notation} & \textbf{Description}\\
		\hline 
		$N$ & the number of tasks\\
		\hline
		$v_j$ & the task with index $j$\\
		\hline
		$q_j(t)$ & the demand of $v_j$ accumulated at time $t$ \\
		\hline
		$ \alpha_j  $ & the inherent increment rate of $v_j$\\
		\hline
		$t_{ij}$ & travel time from $v_i$ to $v_j$ \\
		\hline
		$rob_k$ & the robot with index $k$ \\
		\hline
		$  \beta $ & the ability of a robot\\	
		\hline
		$ at_{jk} $ & the arrival time of $ rob_k $ at $ v_j $\\
		\hline
		$ct_j$ & the completion time of $v_j$ \\
		\hline
		$ x_{ij}^k $ &  1 if  $ rob_k $  goes from $ v_i $ to $ v_j $, and 0 otherwise\\\hline
		\hline
	\end{tabular}
	\vspace{-4mm}
\end{table}

\section{The proposed HDMOEA-$\varepsilon$C algorithm}
The framework of the proposed HDMOEA-$\varepsilon$C method is shown in Algorithm \ref{alg_framework}.
HDMOEA-$\varepsilon$C contains four main components, 1) initialisation, 2) reproduction, 3) matching, and 4) Pareto updating. 
HDMOEA-$\varepsilon$C selects the maximal completion time of all the tasks as the main objective, and converts the number of using robots as the constraints. In the algorithm, $ R $ upper bounds for the number of using robots are maintained for the evolutionary process, which is the same as the population size. At the beginning of HDMOEA-$\varepsilon$C, $ R $ solutions are initialised based on the heuristic method and random method in the designed initialisation method. Then, the rest three components of HDMOEA-$\varepsilon$C are run iteratively until the number of fitness evaluations is large than the maximum number of fitness evaluations. In each generation of the proposed method,  $ I $ solution are  selected to reproduce new solutions $ \textbf{Y} $ firstly. Second, for each solution of the new generated solutions $ \textbf{Y} $, it is matched to a subproblem using the subproblem-to-solution  matching method. Finally, the external archive $ EP $ is updated according to the solutions of the current population.

\begin{algorithm}[tb]
	\footnotesize
	\caption{The proposed HDMOEA-$\epsilon$C method}
	\label{alg_framework}
	\begin{flushleft}
		\textbf{INPUT:} A MO-MPDA instance, related parameters. \\
		\textbf{OUTPUT:} An external archive population $ EP $.
	\end{flushleft}
	\vspace{-2mm}
	\begin{algorithmic}[1]
		\STATE Calculate the upper bound and low bound of the number of robots.
		\STATE Initialise $ R $ evenly spread upper bound vectors,  and initialise the  evolving population $ Pop = \{x^1, \dots, x^R\} $ according to the spread upper bound vectors using Algorithm \ref{alg_init}.
		\STATE Extract nondominated individuals from $ Pop $ denoted  as $ EP $. $ gen  =0, n = N $
		\FOR {$ i =1 $ to $ R $}
		\STATE Set the neighbourhood of the $ i $th subproblem, $ \pi^i = 1$.
		\ENDFOR
		\IF{ $ gen $ is a multiple of $ DRA\_interval $ }
		\STATE Update the indices of the subproblem $ I $ that will be evolved in the next generations.
		\ENDIF
		\WHILE {$ n \leq  NFE$}
		\FOR {$ i \in I $}
				\STATE $P = \left\{ \begin{array}{l}
		B(i),~~~~~~~~~~\text{if}~rand < \delta \\
		\{1,2,...,R\},~\text{otherwise}
		\end{array} \right.$
		\STATE Reproduction (Algorithm \ref{alg_reproduction}): Select parent individuals from $ P $ randomly and apply certain reproduction operator to generate new individuals $ \textbf{Y} $.
		\FOR {$ y \in \textbf{Y} $}
		\STATE If a new individual  $ y  $ is infeasible, repair it.
		\STATE Use the subproblem-to-solution matching method (Algorithm \ref{alg_subproblem}) to find subproblems $ \textbf{K} $ for the new solution $ y $ .
		\STATE Compare $ y $ with solutions of the subproblems $ \textbf{K} $, and update solutions	and neighbouring solutions of the subproblems $ \textbf{K} $.		
		\STATE Update the external archive $ EP $.
		\ENDFOR
		\ENDFOR
		\STATE $ gen = gen + 1 $
		\ENDWHILE		
	\end{algorithmic}
\end{algorithm}

\subsection{Representation and Decoding Strategy}
\subsubsection{Representation}
The explicit representation of the execution plan of all the robots in the MPDA problem is a variable-length sequence of events. To simplify the representation, an implicit representation of a solution for the MO-MPDA problem is adopted in this paper, which is a matrix and shown in \eqref{sol}.  
\begin{equation}
\label{sol}
X = \left[ {\begin{array}{*{20}{c}}
	\pi_{1,[1]} &\pi_{1,[2]} & \cdots &\pi_{1,[N]}\\
	\pi_{2,[1]} &\pi_{2,[2]}& \cdots &\pi_{2,[N]}\\
	\vdots & \vdots & \vdots & \vdots \\
	\pi_{m,[1]} &\pi_{m,[2]} & \cdots &\pi_{m,[N]}
	\end{array}} \right]
\end{equation}
The size of rows of the solution represents the number of robots using in the MO-MPDA problem. Each row of the given representation has $ N $ integral elements which is a permutation of all tasks' indexes.  Similar to the representation in VRPs \cite{wang2018two, long2019hybrid}, the elements of one row indicate the task-executing sequences. For example, if  $ i $th row is $ [2, 3, 1] $,
$ rob_i $ will intend to execute tasks $ v_2 $, $ v_3 $ and $ v_1 $ in order.

\subsubsection{Decoding} 
\begin{figure}[!t]\small
	\centering
	\includegraphics[width=0.95\linewidth]{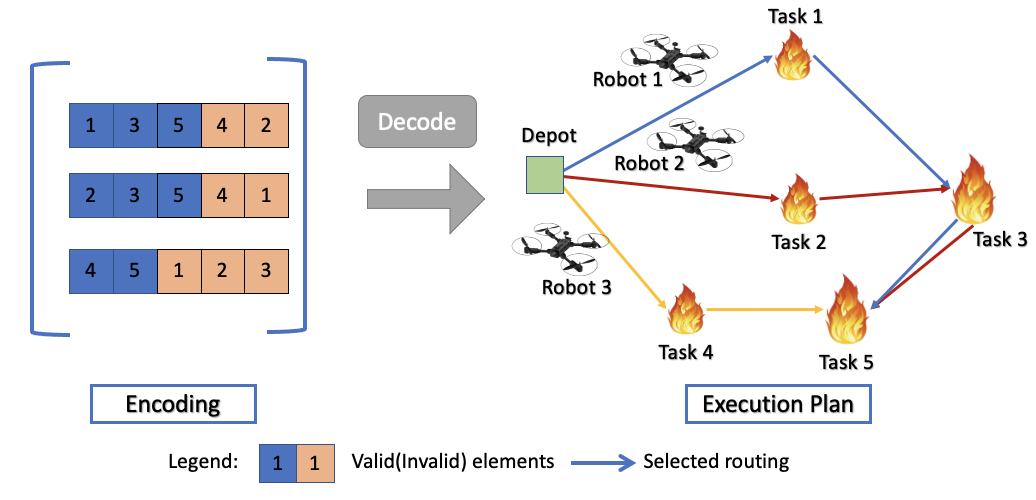}	
	\caption{An example of decoding processes of the MPDA problem with the given encode.}
	\label{mpda_example}
	\vspace{-4mm}
\end{figure}

The decoding method used in this paper adopts the event trigger mechanism. When a robot completed a task, it becomes active and selects its next executing task according to the corresponding encoding. The details of the decoding method can be found in \cite{gao2020ACO,gao2020memetic}. Fig. \ref{mpda_example} shows an example of the decoding process of the MPDA problem with the given encoding.  There are three robots and five tasks in the example, and each robot executes tasks in sequence according to the given encoding. Task $ v_5 $ which is executed by three robots simultaneously is the last completed task. Thus, for the MO-MPDA problem, 
the first objective of the example solution in Fig.~\ref{mpda_example} is the completion time of $ v_5 $, and the second objective is 3.

\textbf{Remark:} A robot will not visit all tasks according to the give encode. There may be some invalid elements in each row of the given encoding.
The number of invalid elements in a row is denoted as $ ie_k $ in this paper. For example, $ ie_3 $ of $ rob_3 $ is 3 in Fig.~\ref{mpda_example}. 
It also should be noticed that many encoding representations have the same objective values since the abilities of robots are the same. For example, swapping the visiting sequences of $ rob_1 $ and $ rob_2 $ does not affect the objective values.

\subsection{The designed Hybrid initialisation mechanism}

\subsubsection{Bound determination}
At the beginning of the designed hybrid initialisation mechanism, the low bound $ \textbf{LBM} $ and upper bound $ \textbf{UBM} $ of the number of used robots are calculated as shown in Eqs. \eqref{lbm} and \eqref{ubm}.
\begin{equation}
\label{lbm}
 \textbf{LBM}  = \lceil \frac{\max_{i \in \{1,\cdots,R\}} \alpha_i }{\beta} \rceil
\end{equation}
\begin{equation}
\label{ubm}
\textbf{UBM} = \sum_{i = 1} ^ N \lceil \frac{\alpha_i}{\beta} \rceil + 1
\end{equation}
where $ \textbf{LBM}  $ represents the minimal number of used robots which ensures every task can be completed, and $ \textbf{UBM} $ represents the maximal number of robots which ensures all the tasks can be executed by one assignment. Meanwhile, a series of scalar upper bounds $ {\bf{\varepsilon }}=({\varepsilon _1},\varepsilon _2,...,{\varepsilon _R}) $ for the number of used robots are generated within an equal space $ \bigtriangleup = 1/(R-1)$. 
Thus, a subproblem $ sp_l $ of HDMOEA-$\varepsilon$C is shown as following.
\begin{equation}
\label{subproblem_eq}
\begin{array}{l}
\begin{array}{*{20}{c}}
{{\mathop{\rm minimize}\nolimits} }&{{f_{main}} = {f_1}({\bf{x}})}
\end{array} \\
{\rm{~~subject~to}}  \left\{ \begin{array}{l}
f_2 \le \varepsilon_l (\textbf{UBM} - \textbf{LBM}) + \textbf{LBM}\\
{\bf{x}} \in \Omega
\end{array} \right.
\end{array}
\end{equation}
It should be noticed that the objective of the robot number is always integer. It is also obvious that the robot number of the optimal solution $ \textbf{x} $  of $ sp_l $ is 
$ \lfloor \varepsilon_l (\textbf{UBM} - \textbf{LBM}) + \textbf{LBM} \rfloor $.

\subsubsection{Population initialisation}
Due to the complex relationships among robots and tasks of the MO-MPDA problem, a purely random initialisation cannot provide solutions of good-quality, even cannot generate feasible solutions.  Conversely, an elaborate heuristic method is competent to find an acceptable solution at a small computational cost.
Incorporating the random and heuristic initialisation methods, a hybrid initialization strategy shown in Algorithm \ref{alg_init} is developed to generate the
initial population for each subproblem in the proposed HDMOEA-$\varepsilon$C method. For each subproblem, its solution is generated according to $ m $ determined using robots (line 3 of Algorithm \ref{alg_init}). For a specific determined number of using $ m $ robots, the heuristic generation method shown in Algorithm \ref{alg_h} is preferred over the random solution generation method. Provided that the heuristic method has generated a solution for a specific $ m $, the rest of the solutions based on   $ m $ robots are generated by the random method. 

\begin{algorithm}[tb]
	\footnotesize
	\caption{Hybrid initialisation strategy}
	\label{alg_init}
	\begin{flushleft}
		\textbf{INPUT:} MO-MPDA instance $ \Xi $. \\
		\textbf{OUTPUT:} Initial population $ Pop $.
	\end{flushleft}
	\vspace{-2mm}
	\begin{algorithmic}[1]
		\STATE $ Pop = \emptyset $, and $ \mathcal{M} $
		\FOR {$ l \rightarrow P $}
		\STATE  Calculate $ \lfloor \varepsilon_l (\textbf{UBM} - \textbf{LBM}) + \textbf{LBM} \rfloor  $ as $ m $.
		\IF {$ m \in \mathcal{M} $}
		\STATE 	$ sol $	= \texttt{RandomGenerate}($ \Xi , m $)
		\ELSE
		\STATE  $ sol $ = \texttt{HeuristicGenerate}($ \Xi , m $), and $ \mathcal{M} $ = $ \mathcal{M} \cup m$ 
		\ENDIF
		\STATE $ Pop = Pop \cup sol$
		\ENDFOR
	\end{algorithmic}
\end{algorithm}	

Many factors including the travelling cost,  increment rates of tasks and the number of using robots contribute to the complexity of the MO-MPDA problem.  A greedy heuristic method considering one factor cannot generate good solutions for different instance with different characteristics. Thus, a comprehensive heuristic method considering the arrival time, completion time of robots and workload balance, is designed in this paper. The details of the designed heuristic method are shown in Algorithm \ref{alg_h}. The heuristic method generates $ 2|\textbf{W}| $ solutions with different weights and sorting rules, the best performance solution $ sol^* $ of $ 2|\textbf{W}| $ generated solutions are complemented and returned as the initial solution for the corresponding subproblem.  The reason for the complementing process is that a robot may not visit all the tasks in the best performance solution $ sol^* $. To keep the same encoding length, the rest tasks which are not assigned to the robot are shuffled and added to the end of the sequence. 
Provided that the sorting rule and weight $ \omega $ are given, the inner loop of the heuristic algorithm adopt the event trigger mechanism. We divide the inner loop generation method into three stages: initialisation stage (line 4), decision-making stage (lines 6-25), and updating stage (line 26).

In the initialisation stage, the solution is set as empty, and the states of robots and tasks are initialised. When robots become active, they calculate the priorities of the candidate tasks in the set $ \mathcal{T}^{(t)}$ according to the perdition arrival and completion time with the given sorting rule and $ \omega $. To normalise the arrival and completion time of different tasks, the priorities of the candidate tasks is obtained by the weighted orders of the sorted arrival time $ \textbf{AT} $ and sorted completion time $ \textbf{CT} $. When the boolean rb is 1, $ \textbf{CT} $ is sorted from small to large. Otherwise, $ \textbf{CT} $ is  sorted from large to small.  It should be noticed that 
{\color{blue} the completion time rules}.
Each active robot selects the task with the highest priority to execute next.  After that, the states of robots and tasks are updated by Algorithm \texttt{StateUpdate}, whose details can be found in \cite{gao2020memetic,gao2020ACO}. Meanwhile, Algorithm \texttt{StateUpdate} returns the new set of active robots for the next round of the decision-making process. 

\begin{algorithm}[tb]
	\footnotesize
	\caption{Heuristic Generation algorithm \texttt{HeuristicGenerate}($ \Xi , m $)}
	\label{alg_h}
	\begin{flushleft}
		\textbf{INPUT:} MO-MPDA instance $ \Xi $, number of using robots $m $. \\
		\textbf{OUTPUT:} Solution $ sol^* $.
	\end{flushleft}
	\vspace{-2mm}
	\begin{algorithmic}[1]
		\STATE A list of solutions $ \textbf{LSOL} $, and generate a series weights $\textbf{W}$ with the same space $ \bigtriangleup $
		\FOR {rb $ \in \{0,1\} $}
		\FOR {$ \omega  \in \textbf{W} $}
		\STATE Initialisation stage of the construction process, $ sol = () $
		\WHILE {There are uncompleted tasks}
		\STATE Get all uncompleted tasks $ \mathcal{T}^{(t)}$, and $ \Pi = \emptyset $.
		\FOR {$ rob_k  \in \mathbf{AR} $}		
		\STATE $ \textbf{AT} = \emptyset $, $ \textbf{CT} = \emptyset $.
\FOR {$ v_i \in \mathcal{T}^{(t)} $}
\STATE Calculate prediction arrival time $ at_ik $, prediction completion time $ ct_ik $, and prediction incremental rate $ \lambda_i (t) $.
\STATE $ \textbf{AT} = \textbf{AT} \cup at_ik$  
\ENDFOR
\STATE Sort $ \textbf{AT}$ from small to large.
\IF{rb == 1}
\STATE Sort $ \textbf{CT} $ from small to large.
\ELSE 
\STATE Sort $ \textbf{CT} $ from large to small.
\ENDIF 
\FOR {$  v_i \in \mathcal{T}^{(t)} $}
\STATE Get the order of $ v_i $ in $ \textbf{AT} $  and $ \textbf{CT}$ respectively, denoted as $ \textbf{AT}_{[v_i]} $ and $ \textbf{CT}_{[v_i]} $.
\STATE $ p_i = \omega \textbf{AT}_{[v_i]}  +   (1- \omega) \textbf{CT}_{[v_i]} $.
\ENDFOR
\STATE  Assign $ v_k^* = \arg \max_{i= 1}^{|\varOmega|}  p_i $ to $  rob_k $.
\ENDFOR
\STATE Set $ \Pi  = {[rob_k, v^*_k] | rob_k \in \mathbf{AR} } $.
\STATE [$ \mathbf{AR}, sol  $] = \texttt{StateUpdate}($ \Pi , sol$)
		\ENDWHILE 
		\STATE $ \textbf{LSOL} = \textbf{LSOL} \cup sol $
		\ENDFOR
		\ENDFOR
		\STATE $ sol^* = \arg \max_{i= 1}^{|\textbf{LSOL}|}  f_1(sol_i) $, and complement $ sol^*$.
	\end{algorithmic}
\end{algorithm}

\subsection{Reproduction strategy}
The reproduction strategy is an important aspect of MOEAs, which generates new offspring solutions through transferring information among the parent solutions.
And the reproduction process not only is expected to generate diversified solutions,  but also is able to propagate excellent information from parents \cite{holland1992adaptation}. The representation of the MO-MPDA problem is multiple permutations of the tasks' indexes,  and previous researchers have developed a large number of crossover operators and mutation operators \cite{potvin1996vehicle,jozefowiez2009evolutionary}. Thus,   we developed the reproduction strategy based on the classical crossover operator (partially matched crossover) and mutation operator (swap mutation) \cite{karakativc2015survey}, which is shown in Algorithm \ref{alg_reproduction}. In the proposed reproduction strategy, there are two generation operators. When a sampled value less than $ \delta_g $,  the new offspring solutions are provided by the designed crossover operator. Otherwise, the new offspring solutions are generated by the designed mutation operator.

Since the robots in the MO-MPDA problem are homogeneous, the permutation encoding has great redundancy. 
At the beginning of the designed crossover operator, permutations in each parent solution are sorted according to the number of invalid elements and task indexes to address the redundancy issue. Then, the two situations are distinguished. The first situation is that parent solutions $ X_a $ and $ X_b $ have the same number of using robots. For each permutation of the parent solutions, the partially matched crossover generates offspring permutations to construct the new offspring solutions $ \textbf{Y} $.
When the parent solutions $ X_a $ and $ X_b $ have different numbers of using robots, the new offspring solutions $ \textbf{Y} $ are constructed by three parts.
The first part of the new offspring solutions is similar to the previous situation, and it includes two generated solutions based on the partially matched crossover. 
The second part is based on the second part, and it includes up to 10 new solutions, which is constructed by selecting $ f_2(Y_a) $ permutations from the solution with a large number of robots  $ f_2(Y_a) $ to construct new solutions. The last part is generated by inheriting the alleles from the two parent individuals with no implicit mutations.

\begin{algorithm}[tb]
	\footnotesize
	\caption{Reproduction strategy \texttt{Reproduce}($ X_a $, $ X_b$)}
	\label{alg_reproduction}
	\begin{flushleft}
		\textbf{INPUT:} . Parents solution $ X_a $ and $ X_b $ \\
		\textbf{OUTPUT:} New offspring solutions $ \textbf{Y} $.
	\end{flushleft}
	\vspace{-2mm}
	\begin{algorithmic}[1]		
		\IF { $ rand  < \delta_g $ }
		\STATE Sort  $ X_a $ and  $ X_b $ based on the number of invalid elements and task indexes from small to large.
		\IF {$ X_a $ and $ X_b $ have the same number of using robots}		
		\FOR { $ i \rightarrow f_2(X_a) $}
		\STATE The partially matched crossover operates $ X_{a,i} $ and $ X_{b,i} $ to generate offspring permutations $ Y_{a,i} $ and $ Y_{b,i} $.
		\ENDFOR
		\STATE  $ \textbf{Y} =  \{Y_a, Y_b\}$.  
		\ELSE		
		\FOR{ $ i \rightarrow \min\{f_2(X_a), f_2(X_b)\} $}
		\STATE The partially matched crossover operates $ X_{a,i} $ and $ X_{b,i} $ to generate offspring permutations $ Y_{a,i} $ and $ Y_{b,i} $.
		\ENDFOR 
		\STATE  $ \textbf{Y} =  \{Y_a, Y_b\}$.  
		\IF {$ f_2(Y_a) > f_2(Y_b) $}
		\STATE Swap $ Y_a $ and $ Y_b $.
		\ENDIF 
		\FOR {$ i \rightarrow \min\{\tbinom{f_2(Y_a)}{f_2(Y_b)}, 10 \} $}
		\STATE Select $ f_2(Y_a) $ permutations randomly of  $ Y_b $ to construct $ Y_c $. 
		\STATE $ \textbf{Y} =  \textbf{Y} \cup  Y_c$
		\ENDFOR 
		\FOR{ $ i \rightarrow \min\{f_2(X_a), f_2(X_b)\} $}
	\STATE Swap $ X_{a,i} $ and $ X_{b,i} $ to generate $ Y_{a,i} $ and $ Y_{b,i} $.
	\ENDFOR 
	\STATE  $ \textbf{Y} =  \textbf{Y} \cup \{Y_a, Y_b\}$.  		
		\ENDIF 
		\ELSE
		\STATE Swap mutation operates each permutation of  $ X_a $ and $ X_b $ to generate offspring $ Y_{a} $ and $ Y_{b} $.
		\STATE  $ \textbf{Y} =  \{Y_a, Y_b\}$.  
		\ENDIF
	\end{algorithmic}
\end{algorithm}

\subsection{Subproblem-to-Solution}
When a new solution is generated, it is necessary to determine which subproblem is suitable for the solution. 
To make the best use of information about the new generated solutions, we develop a subproblem-to-solution matching procedure, which is shown in Algorithm \ref{alg_subproblem}. In the matching procedure, the subproblem that does not violate constraints and has the maximal makespan objective is selected for the newly generated solution.  Since the feasibility rule is adopted to handle constrained subproblems, this procedure is good for convergence. The solution-to-subproblem matching procedure and the subproblem-to-solution matching procedure consider diversity and convergence, respectively.

\begin{algorithm}[tb]
	\footnotesize
	\caption{Subproblem-to-Solution Matching}
	\label{alg_subproblem}
	\begin{flushleft}
		\textbf{INPUT:} $N$ subproblems ${{\bf{\varepsilon }}^{1}}{\bf{,}}{{\bf{\varepsilon }}^{2}}...{\bf{,}}{{\bf{\varepsilon }}^{N}}$ and new generated solution ${\bf{y}}$.
		\textbf{OUTPUT:} The index of the selected subproblem  $ k $.
	\end{flushleft}
	\vspace{-2mm}
	\begin{algorithmic}[1]		
		\FOR {$ i  \rightarrow R$}
		\STATE $C{V^i} = \min (\varepsilon_i - \frac{{f_2(\bf{y}) - \textbf{LBM}}}{{\textbf{UBM} - \textbf{LBM}}} ,0)$
		\IF {$C{V^i} \neq 0$}
		\STATE $ C{V^i} = f_1(sol_i) $
		\ENDIF 
		\ENDFOR
		\STATE $ k = \arg \max_{1,...,R}(CV^1,...,CV^R) $.
	\end{algorithmic}
\end{algorithm}	

\subsection{Discussion}

\subsubsection{Time Complexity Analysis} 
The proposed HDMOEA-$\epsilon$C method is used to optimise the MO-MPDA problem in which the number of  tasks is $ N $, and the upper bound of the robot number is 
$ \textbf{UBM} $.  In each iteration, there are $ R $ solution in the population to evolve. Table \ref{time_complexity} shows the time complexity of HDMOEA-$\epsilon$C. 
In the initialisation process, the time complexity mainly consists of the random and heuristic generation methods.
In the evolutionary process, the time complexity mainly lies in the dynamic resource allocation, reproduction, subproblem-to-solution, updating $ EP $, and decoding process.
In summary, the time complexity  of HDMOEA-$\epsilon$C is $  {\rm{O}} (R (\textbf{UBM}\times N (N + \textbf{UBM}))  + R ^2)$.

\begin{table}[tb]
	\centering
	\caption{Time complexity of HDMOEA-$\epsilon$C.}
	\begin{tabular}{c|c}
		\hline
		\hline
		Parameter & Value \\
		\hline
		Random solution generation & $ {\rm{O}}(\textbf{UBM} \times  N) $ \\ \hline
		Heuristic solution generation & $ {\rm{O}}(\textbf{UBM}\times N (N + \textbf{UBM}))$  \\ \hline
		Dynamic resource allocation & $ {\rm{O}}(R) $\\ \hline 
		Reproduction process & $ {\rm{O}}(R\times N \times \textbf{UBM}) $ \\ \hline  
		Subproblem-to-solution matching & $ {\rm{O}}(R) $ \\ \hline 
		Update $ EP $  &  $ {\rm{O}}(R^2) $\\ \hline 
		Decoding process & $ {\rm{O}}(\textbf{UBM}\times N (N + \textbf{UBM}))$  \\ \hline 		
		\hline
	\end{tabular}
	\vspace{-4mm}
	\label{time_complexity}
\end{table}

\section{DESIGN OF EXPERIMENTS}
\subsection{Data Set} 
There is not existing benchmark set for the MO-MPDA problem. To fair compare performances of different algorithms on the MO-MPDA problem, a comprehensive MO-MPDA benchmark set is developed in this paper. Following the characteristics of MPDA problem and VRPs \cite{gao2020ACO,gao2020memetic,uchoa2017new}, we transformed the single-objective MPDA benchmark set to a MO-MDPA benchmark set. The settings of instances of the MO-MPDA problem including the number of tasks, task initial demands and inherent increment rates follows the settings of the most comprehensive MPDA benchmark set in \cite{gao2020ACO}.  
The abilities of each robots of the designed MO-MPDA is set as the mean abilities of all robots in the MPDA benchmark set in \cite{gao2020ACO}.
Table  \ref{ins} shows the full details of all the designed instances, and the instances are named by the number of tasks, the position of tasks, the abilities of robots, and the inherent increment rate of tasks \footnote[1]{We will publish the  designed benchmark set after the paper is accepted.} . 

\begin{table}[tb]
	\caption{Main characteristics of forty-five designed instances.
	}
	\label{ins}
	\centering
		\setlength\tabcolsep{3pt}
	
	\begin{threeparttable}
		\begin{tabular}{|c| c| c| c |c |c| c| c| c |c |c| c| }
			\hline
			ID & name & $ N $  &PR & PT & $  \beta $  & $ \alpha $  \\
			\hline			
1  & 5\_C\_CL\_2.86 & 5 &C &CL &[0.02, 0.05] &[0.05, 0.15]\\ \hline 
2  & 5\_EC\_RCL\_2.86 & 5 &EC &RCL &[0.02, 0.05] &[0.05, 0.15]\\ \hline 
3  & 10\_C\_R\_0.54 & 10 &C &R &[0.05, 0.08] &[0.02, 0.05]\\ \hline 
4  & 10\_EC\_CL\_0.54 & 10 &EC &CL &[0.05, 0.08] &[0.02, 0.05]\\ \hline 
5  & 10\_EC\_RCL\_1.86 & 10 &EC &RCL &[0.02, 0.05] &[0.05, 0.08]\\ \hline 
6  & 10\_C\_R\_4.29 & 10 &C &R &[0.02, 0.05] &[0.1, 0.2]\\ \hline 
7  & 10\_C\_RCL\_1.86 & 10 &C &RCL &[0.02, 0.05] &[0.05, 0.08]\\ \hline 
8  & 10\_C\_CL\_1.0 & 10 &C &CL &[0.05, 0.08] &[0.05, 0.08]\\ \hline 
9  & 10\_EC\_RCL\_1.86A & 10 &EC &RCL &[0.02, 0.05] &[0.05, 0.08]\\ \hline 
10  & 10\_C\_CL\_1.86 & 10 &C &CL &[0.02, 0.05] &[0.05, 0.08]\\ \hline 
11  & 10\_C\_R\_1.86 & 10 &C &R &[0.02, 0.05] &[0.05, 0.08]\\ \hline 
12  & 10\_C\_CL\_4.29 & 10 &C &CL &[0.02, 0.05] &[0.1, 0.2]\\ \hline 
13  & 10\_EC\_R\_2.86 & 10 &EC &R &[0.02, 0.05] &[0.05, 0.15]\\ \hline 
14  & 15\_EC\_CL\_1.0 & 15 &EC &CL &[0.05, 0.08] &[0.05, 0.08]\\ \hline 
15  & 15\_EC\_CL\_1.0A & 15 &EC &CL &[0.02, 0.05] &[0.02, 0.05]\\ \hline 
16  & 15\_EC\_CL\_1.86 & 15 &EC &CL &[0.02, 0.05] &[0.05, 0.08]\\ \hline 
17  & 15\_C\_RCL\_4.29 & 15 &C &RCL &[0.02, 0.05] &[0.1, 0.2]\\ \hline 
18  & 20\_EC\_R\_1.0 & 20 &EC &R &[0.05, 0.08] &[0.05, 0.08]\\ \hline 
19  & 20\_C\_R\_2.86 & 20 &C &R &[0.02, 0.05] &[0.05, 0.15]\\ \hline 
20  & 20\_EC\_CL\_0.54 & 20 &EC &CL &[0.05, 0.08] &[0.02, 0.05]\\ \hline 
21  & 20\_C\_R\_4.29 & 20 &C &R &[0.02, 0.05] &[0.1, 0.2]\\ \hline 
22  & 20\_EC\_R\_0.54 & 20 &EC &R &[0.05, 0.08] &[0.02, 0.05]\\ \hline 
23  & 20\_EC\_RCL\_1.0 & 20 &EC &RCL &[0.05, 0.08] &[0.05, 0.08]\\ \hline 
24  & 20\_EC\_RCL\_2.86 & 20 &EC &RCL &[0.02, 0.05] &[0.05, 0.15]\\ \hline 
25  & 20\_C\_RCL\_4.29 & 20 &C &RCL &[0.02, 0.05] &[0.1, 0.2]\\ \hline 
26  & 30\_C\_CL\_1.0 & 30 &C &CL &[0.02, 0.05] &[0.02, 0.05]\\ \hline 
27  & 30\_EC\_R\_1.0 & 30 &EC &R &[0.02, 0.05] &[0.02, 0.05]\\ \hline 
28  & 30\_C\_RCL\_1.86 & 30 &C &RCL &[0.02, 0.05] &[0.05, 0.08]\\ \hline 
29  & 30\_EC\_CL\_2.86 & 30 &EC &CL &[0.02, 0.05] &[0.05, 0.15]\\ \hline 
30  & 30\_C\_R\_4.29 & 30 &C &R &[0.02, 0.05] &[0.1, 0.2]\\ \hline 
31  & 40\_C\_R\_0.54 & 40 &C &R &[0.05, 0.08] &[0.02, 0.05]\\ \hline 
32  & 40\_C\_CL\_1.86 & 40 &C &CL &[0.02, 0.05] &[0.05, 0.08]\\ \hline 
33  & 40\_EC\_CL\_1.0 & 40 &EC &CL &[0.02, 0.05] &[0.02, 0.05]\\ \hline 
34  & 40\_EC\_R\_1.86 & 40 &EC &R &[0.02, 0.05] &[0.05, 0.08]\\ \hline 
35  & 40\_EC\_R\_4.29 & 40 &EC &R &[0.02, 0.05] &[0.1, 0.2]\\ \hline 
36  & 60\_C\_CL\_1.0 & 60 &C &CL &[0.02, 0.05] &[0.02, 0.05]\\ \hline 
37  & 60\_C\_R\_0.54 & 60 &C &R &[0.05, 0.08] &[0.02, 0.05]\\ \hline 
38  & 60\_C\_R\_1.0 & 60 &C &R &[0.05, 0.08] &[0.05, 0.08]\\ \hline 
39  & 60\_C\_CL\_1.0A & 60 &C &CL &[0.02, 0.05] &[0.02, 0.05]\\ \hline 
40  & 60\_C\_R\_1.0A & 60 &C &R &[0.02, 0.05] &[0.02, 0.05]\\ \hline 
41  & 60\_EC\_R\_1.0 & 60 &EC &R &[0.02, 0.05] &[0.02, 0.05]\\ \hline 
42  & 80\_EC\_CL\_1.0 & 80 &EC &CL &[0.02, 0.05] &[0.02, 0.05]\\ \hline 
43  & 80\_EC\_CL\_0.54 & 80 &EC &CL &[0.05, 0.08] &[0.02, 0.05]\\ \hline 
44  & 80\_EC\_R\_0.54 & 80 &EC &R &[0.05, 0.08] &[0.02, 0.05]\\ \hline 
45  & 120\_EC\_RCL\_1.0 & 120 &EC &RCL &[0.05, 0.08] &[0.05, 0.08]\\ \hline 
		\end{tabular}
\begin{tablenotes}
\scriptsize  
\item[*] PR means positions of robots, PT means positions of tasks, C represents central, E represent  Eccentric, R represents random, CL represents Clustered, and RC represents Random-Clustered.
\end{tablenotes}
\end{threeparttable}
\end{table}

\subsection{Competitor Algorithms}
Since the MO-MPDA problem is a novel problem and no existing algorithms can be directly applied for comparison, we compare HDMOEA-$\varepsilon$C with the following baseline algorithms: 

\begin{itemize}
	\item NSGA-II \cite{deb2002fast}. It is one of the most popular MOEAs, selects individuals according to Pareto dominance relation and propagates offspring in an iterative way. The main feature of this algorithm is that it adopts the elitist nondominated sorting with the crowding distance as a ranking criterion.
	\item MOEA/D \cite{zhang2007moea} and  MOEA/D-DRA \cite{zhang2009performance}. They are both developed according to the decomposition strategy and outperform a number of popular algorithms.
	\item MOEA/D-VLP  \cite{li2019variable}. The MO-MPDA problem is a variable-length multi-objective optimisation problem. MOEA/D-VLP  is a state-of-the-art algorithm for the variable-length optimisation problem. In MOEA/D-VLP, it decomposes a multi-objective problem in terms of the penalty boundary intersection search directions and the dimensionality of variables.
	\item MOEA/D-ACO \cite{ke2013moea}. The adaptive ACO method proposed in \cite{gao2020ACO} is a state-of-the-art algorithm for the single-objective MPDA problem.  Following the ideas in \cite{ke2013moea}, we combine the adaptive ACO method and MOEA/D to address the MO-MPDA problem.	
\end{itemize}

\subsection{Performance metrics and parameter settings}
\subsubsection{Performance metrics}
Three commonly used performance metrics, i.e., inverted generational distance (IGD) \cite{zhou2005model} and hypervolume (HV) \cite{zitzler1999multiobjective} are employed to evaluate the performance of all compared algorithms in this paper. IGD and HV assess the quality of a nondominated set in terms of  convergence and diversity, and their definitions are shown in Eqs. \eqref{m_IGD} and \eqref{m_HV}.
\begin{equation}
\label{m_IGD}
IGD({P^*},P) = \frac{{\sum\nolimits_{{x^*} \in {P^*}} {d({x^*},P)} }}{{|{P^*}|}}
\end{equation}
\begin{equation}
\label{m_HV}
HV(P,{\bf{r}}) = VOL(\bigcup\limits_{x \in P} {[{f_1}(x),{r_1}] \times ... \times [{f_m}(x),{r_m}]} )
\end{equation}
where $ P $ represents the solution set,  $ \bf{r} $ represents the reference point, and  $ {P^*} $ represents the true pareto front. However, the ture pareto front of the MO-MPDA problem is very difficult to obtained due to the complexity of the problem. In this paper, we approximate the true pareto front by selecting non-dominated solutions from all the compared and designed algorithms \cite{ishibuchi2014difficulties}. The IGD and HV  values are calculated based on the normalised objectives of which the range is [0,1], the normalisation methods are shown in Eqs. \eqref{n_ob1} and \eqref{n_ob2}.
\begin{equation}
\label{n_ob1}
ob1_n = \frac{\ln(\max_{i \in \{1,\cdots,N\}} ct_i) - 2\ln(10)}{6\ln(10) - 2\ln(10)}
\end{equation} 
\begin{equation}
\label{n_ob2}
ob2_n = \frac{m -  \textbf{LBM}}{\textbf{UBM} - \textbf{LBM}}
\end{equation}
where $ \max_{i \in \{1,\cdots,N\}} ct_i $  ($ m $) represents the maximal completion (number of robots) objective value, and $ ob1_n $ ($ ob2_n $) represents the first (second) normalised objective value.  The point (1.1, 1.1) is used as the reference point in this paper.

\subsubsection{Parameter settings}
All the competitor algorithms and HDMOEA-$\epsilon$C for the MO-MPDA problem are implemented based on a Python  evolutionary computation framework \cite{DEAP_JMLR2012} to keep fair comparisons.  The parameter settings of the proposed DMOEA-$\epsilon$C method used in the rest of this paper follow the conventional settings \cite{zhang2007moea,chen2017dmoea,li2020dmaoea}. The detailed parameter setting are shown Table \ref{paras}.
For each instance, all the algorithms was run 20 times independently.

\begin{table}[tb]
	\centering
	\caption{Parameter settings of DMOEA-$\epsilon$C.}
	\begin{tabular}{c|c}
		\hline
		\hline
		Parameter & Value \\
		\hline
		Pop Size $N$ & 100\\ \hline 
		Neighbourhood size $ T $ & $ \lfloor 0.1 N \rfloor $ \\ \hline
		Probability  $ \delta $ & 0.9 \\ \hline
		$ DRA\_interval $ & 50 \\ \hline
		Selected subproblem size $ |I| $ & $ \lfloor N/ 5 \rfloor - 2 $ \\ \hline
		Stop criterion & $ NFE $ =  50000 \\ \hline
		Weight size $ |\textbf{W}| $ & 11 \\ \hline
		Mutation probability $ \delta_g $ & 0.9 \\ \hline 
		\hline
	\end{tabular}
	\vspace{-4mm}
	\label{paras}
\end{table}

\section{RESULTS AND DISCUSSIONS}
First, HDMOEA-$\epsilon$C is compared with the state-of-the-art MOEAs on the benchmark set of the MO-MPDA problem. Then, further analyses of the designed hybrid initialisation and reproduction strategies are made to find the reason why HDMOEA-$\epsilon$C is effective on the MO-MPDA problem. Wilcoxon rank-sum test with a $ 5\% $ significance level  and Bonferroni correction  are used to verify in the rest experiments.

the performance of the proposed algorithm.

\subsection{Comparisons With State-of-the-Art Algorithms}
And their results compared using the Wilcoxon ran-sum test with a $ 5\% $ significance level

\subsection{Effectiveness of the hybrid initialisation strategy}

\subsection{Effectiveness of the reproduction strategy}

\section{Conclusion}
The goal of this paper was to address an emerging and novel MO-MPDA problem from the real-world applications. Due to the complex dependencies among robots and tasks,
the redundant encoding, and variable-size decision space, the MO-MPDA problem is a very challenging problem.
Finally, the goal  of this paper has been successfully achieved by proposing  a MO-MPDA model and designing an elaborate HDMOEA-$\epsilon$C method.
Specifically, the two novel strategies of HDMOEA-$\epsilon$C, the hybrid initialisation and the reproduction strategies, have shown great effectiveness and efficiency, respectively.

\ifCLASSOPTIONcaptionsoff
  \newpage
\fi



\bibliographystyle{IEEEtran}
\bibliography{MOMPDA}
%

%





\end{document}


%
\title{Automated Coordination Strategy Design using Genetic Programming  for Dynamic Multi-Point Dynamic Aggregation}
%
%
%
\author{
	Guanqiang~Gao,
Yi~Mei,~\IEEEmembership{Senior Member,~IEEE,}
~Bin~Xin,~\IEEEmembership{Member,~IEEE, }
Ya-hui~Jia,~\IEEEmembership{Member,~IEEE,}
and Will~N.~Browne,~\IEEEmembership{Member,~IEEE}
}

\maketitle

\IEEEpeerreviewmaketitle

\begin{table}[!h]
	\caption{Main characteristics of forty-five designed scenarios.}
	\label{t-scenarios}
	\centering
	\footnotesize
	\begin{tabular}{|c| c| c| c |c |c| c| c| c |c |c| c|  c| c| c|  c| c| c|  c| c|}
		\hline
		Name & $ M $ & $ N $ &$ \mu $ & $ \gamma $ & $ \rho $\\
		\hline
		\hline
		r10t200\_L\_S &10 & 200 &$N(0.035,0.0175) $ &[0.01,0.02] &0.4\\ \hline 
		r10t200\_S\_S &10 & 200 &$N(0.035,0.0035) $ &[0.01,0.02] &0.4\\ \hline 
		r10t200\_L\_M &10 & 200 &$N(0.035,0.0175) $ &[0.02,0.05] &0.65\\ \hline 
		r10t200\_S\_M &10 & 200 &$N(0.035,0.0035) $ &[0.02,0.05] &0.65\\ \hline 
		r10t200\_L\_L &10 & 200 &$N(0.035,0.0175) $ &[0.05,0.08] &0.9\\ \hline 
		r10t200\_S\_L &10 & 200 &$N(0.035,0.0035) $ &[0.05,0.08] &0.9\\ \hline 
		r10t300\_L\_S &10 & 300 &$N(0.035,0.0175) $ &[0.01,0.02] &0.4\\ \hline 
		r10t300\_S\_S &10 & 300 &$N(0.035,0.0035) $ &[0.01,0.02] &0.4\\ \hline 
		r10t300\_L\_M &10 & 300 &$N(0.035,0.0175) $ &[0.02,0.05] &0.65\\ \hline 
		r10t300\_S\_M &10 & 300 &$N(0.035,0.0035) $ &[0.02,0.05] &0.65\\ \hline 
		r10t300\_L\_L &10 & 300 &$N(0.035,0.0175) $ &[0.05,0.08] &0.9\\ \hline 
		r10t300\_S\_L &10 & 300 &$N(0.035,0.0035) $ &[0.05,0.08] &0.9\\ \hline 
		r10t500\_L\_S &10 & 500 &$N(0.035,0.0175) $ &[0.01,0.02] &0.4\\ \hline 
		r10t500\_S\_S &10 & 500 &$N(0.035,0.0035) $ &[0.01,0.02] &0.4\\ \hline 
		r10t500\_L\_M &10 & 500 &$N(0.035,0.0175) $ &[0.02,0.05] &0.65\\ \hline 
		r10t500\_S\_M &10 & 500 &$N(0.035,0.0035) $ &[0.02,0.05] &0.65\\ \hline 
		r10t500\_L\_L &10 & 500 &$N(0.035,0.0175) $ &[0.05,0.08] &0.9\\ \hline 
		r10t500\_S\_L &10 & 500 &$N(0.035,0.0035) $ &[0.05,0.08] &0.9\\ \hline 
		r15t200\_L\_S &15 & 200 &$N(0.035,0.0175) $ &[0.01,0.02] &0.25\\ \hline 
		r15t200\_S\_S &15 & 200 &$N(0.035,0.0035) $ &[0.01,0.02] &0.25\\ \hline 
		r15t200\_L\_M &15 & 200 &$N(0.035,0.0175) $ &[0.02,0.05] &0.5\\ \hline 
		r15t200\_S\_M &15 & 200 &$N(0.035,0.0035) $ &[0.02,0.05] &0.5\\ \hline 
		r15t200\_L\_L &15 & 200 &$N(0.035,0.0175) $ &[0.05,0.08] &0.7\\ \hline 
		r15t200\_S\_L &15 & 200 &$N(0.035,0.0035) $ &[0.05,0.08] &0.7\\ \hline 
		r15t300\_L\_S &15 & 300 &$N(0.035,0.0175) $ &[0.01,0.02] &0.25\\ \hline 
		r15t300\_S\_S &15 & 300 &$N(0.035,0.0035) $ &[0.01,0.02] &0.25\\ \hline 
		r15t300\_L\_M &15 & 300 &$N(0.035,0.0175) $ &[0.02,0.05] &0.5\\ \hline 
		r15t300\_S\_M &15 & 300 &$N(0.035,0.0035) $ &[0.02,0.05] &0.5\\ \hline 
		r15t300\_L\_L &15 & 300 &$N(0.035,0.0175) $ &[0.05,0.08] &0.7\\ \hline 
		r15t300\_S\_L &15 & 300 &$N(0.035,0.0035) $ &[0.05,0.08] &0.7\\ \hline 
		r15t500\_L\_S &15 & 500 &$N(0.035,0.0175) $ &[0.01,0.02] &0.25\\ \hline 
		r15t500\_S\_S &15 & 500 &$N(0.035,0.0035) $ &[0.01,0.02] &0.25\\ \hline 
		r15t500\_L\_M &15 & 500 &$N(0.035,0.0175) $ &[0.02,0.05] &0.5\\ \hline 
		r15t500\_S\_M &15 & 500 &$N(0.035,0.0035) $ &[0.02,0.05] &0.5\\ \hline 
		r15t500\_L\_L &15 & 500 &$N(0.035,0.0175) $ &[0.05,0.08] &0.7\\ \hline 
		r15t500\_S\_L &15 & 500 &$N(0.035,0.0035) $ &[0.05,0.08] &0.7\\ \hline 
		r20t200\_S\_S &20 & 200 &$N(0.035,0.0035) $ &[0.01,0.02] &0.15\\ \hline 
		r20t200\_S\_M &20 & 200 &$N(0.035,0.0035) $ &[0.02,0.05] &0.3\\ \hline 
		r20t200\_S\_L &20 & 200 &$N(0.035,0.0035) $ &[0.05,0.08] &0.6\\ \hline 
		r20t300\_S\_S &20 & 300 &$N(0.035,0.0035) $ &[0.01,0.02] &0.15\\ \hline 
		r20t300\_S\_M &20 & 300 &$N(0.035,0.0035) $ &[0.02,0.05] &0.3\\ \hline 
		r20t300\_S\_L &20 & 300 &$N(0.035,0.0035) $ &[0.05,0.08] &0.6\\ \hline 
		r20t500\_S\_S &20 & 500 &$N(0.035,0.0035) $ &[0.01,0.02] &0.15\\ \hline 
		r20t500\_S\_M &20 & 500 &$N(0.035,0.0035) $ &[0.02,0.05] &0.3\\ \hline 
		r20t500\_S\_L &20 & 500 &$N(0.035,0.0035) $ &[0.05,0.08] &0.6\\ \hline 		
	\end{tabular}
\end{table}

\section{Parameter Sensitivity Analysis on $ \phi $ and $\omega$}

The size $ \phi $ of the queue for recording the history execution period   and the emergent threshold $ \omega $ are two new  parameters in the GPHH method for DMPDA.  If they are set to very small values, the algorithm will focus on gathering robots together to execute task one by one.  On the other hand, if their values are large, 
the algorithm has a poor ability to deal with the emergent tasks.
Here, three scenarios with different scales and characteristics,  r10t300\_L\_S, r15t500\_L\_M,  and r20t300\_S\_L are used to find how these parameters affect the performance. 
$ \phi $ is set to four values $ \{5, 10, 20, 100\} $, $ \omega $ is set to four values \{$ 10, 20,  100, 1000$\}, and other settings are kept unchanged. 
Different parameters on each instance are tested 20 times to get statistical results which are shown in Table \ref{res_para}.

From the table, it can be found that the performance of GPHH tends to be worse with the increase of $ \phi $. 
In the designed emergent task handling mechanism, the opportunity of triggering the handling mechanism decreases with $ \phi $. The reason is that
a large size of the queue $ \phi $ leads a late time for triggering the handling mechanism and a large threshold which is used to distinguish the emergent tasks and normal tasks. Thus, a large $ \phi  $ is not suitable for the proposed GPHH.
From the table, it can also be found that the performance of GPHH improves with the increase of $ \omega$  from $ 10$ to $20 $. However, when $ \omega $ increases to $ 100$ and $ 1000 $, the performance becomes worse. When $ \omega $ is a very small value, the threshold used for distinguishing the emergent and normal tasks is so small that majority tasks are classified into emergent tasks. Robots are forced to gather together to execute majority tasks. When $ \omega $ is a very large value, the threshold is so large that emergent tasks are difficult to be found. The accumulated demands of the emergent tasks lead to the poor performance of the whole multi-robot system.
Besides,  when $ \phi $ and $  \omega $ are too small or too large, some RCSs obtained by the GPHH method cannot generate feasible solutions for all tested instances in 20 runs. From the aspect of the instance, it can be found when increment rates of tasks are small, the performance of GPHH is not sensitive to the parameter of $ \phi $ and $ \omega $.  In summary,  $\phi = 20 $ and $ \omega = 20 $, which maintain a good performance for different scenarios, are adopted in this paper.

\begin{table*}[!h]
	\caption{Statistical results about parameter $ \phi  $ and $ \omega $.
	}
	\footnotesize
	\label{res_para}
	\centering
	\begin{threeparttable} 
		\begin{tabular}{|c| c| c| c |c |c| c| c| c |c |c|  c|}
			\hline
			\multirow{2}{*}{\shortstack{$ \phi $}}  &\multirow{2}{*}{\shortstack{$ \omega $}} &\multicolumn{2}{c|}{r10t300\_L\_S}   &\multicolumn{2}{c|}{r15t500\_L\_M} &\multicolumn{2}{c|}{r20t300\_S\_L}  \\
			\cline{3-8}
			&	&Mean & Std  &Mean & Std  &Mean & Std \\			
			\hline
			5 &10 &1.06E+4 &9.0E+2 & *& * &7.93E+4 &3.0E+4\\ \hline 
			5 &20 &1.07E+4 &1.0E+3 &4.17E+4 &1.1E+4 &7.58E+4 &2.3E+4\\ \hline 
			5 &100 &1.06E+4 &9.0E+2 &4.21E+4 &9.3E+3 &3.11E+7 &1.4E+8\\ \hline 
			5 &1000 &1.05E+4 &6.7E+2 &4.22E+4 &1.6E+4 &8.12E+4 &3.4E+4\\ \hline 
			10 &10 &1.05E+4 &6.4E+2 &1.23E+5 &3.3E+5 &6.27E+4 &1.0E+4\\ \hline 
			10 &20 &1.07E+4 &6.9E+2 &4.71E+4 &1.4E+4 &6.89E+4 &1.3E+4\\ \hline 
			10 &100 &1.10E+4 &1.1E+3 &4.90E+4 &1.8E+4 &6.85E+4 &1.7E+4\\ \hline 
			10 &1000 &1.09E+4 &1.2E+3 &4.96E+4 &2.0E+4 &2.22E+12 &9.7E+12\\ \hline 
			20 &10 &1.08E+4 &1.4E+3 &4.34E+4 &1.1E+4 &1.61E+12 &7.0E+12\\ \hline 
			20 &20 &1.05E+4 &7.2E+2 &4.38E+4 &1.3E+4 &6.54E+4 &1.0E+4\\ \hline 
			20 &100 &1.07E+4 &7.4E+2 &4.22E+4 &1.3E+4 &7.55E+4 &2.4E+4\\ \hline 
			20 &1000 &1.08E+4 &1.2E+3 &4.66E+4 &1.7E+4 &8.13E+4 &3.3E+4\\ \hline 
			100 &10 &1.03E+4 &6.2E+2 &4.73E+4 &1.7E+4 &7.64E+4 &2.2E+4\\ \hline 
			100 &20 &1.11E+4 &1.3E+3 &4.38E+4 &1.5E+4 &8.23E+4 &2.7E+4\\ \hline 
			100 &100 &1.04E+4 &6.7E+2 &4.24E+4 &9.8E+3 & *& *\\ \hline 
			100 &1000 &1.03E+4 &5.4E+2 &4.03E+4 &9.0E+3 & *& *\\ \hline 
		\end{tabular}
		\begin{tablenotes}
			\scriptsize  
			\item[] * represents that the obtained RCS by the GPHH method cannot generate a feasible solution in the tested process for all 20 runs.
		\end{tablenotes}
	\end{threeparttable}
\end{table*}
